%% file: root.tex
\pdfoutput=1

\documentclass[letterpaper, 10 pt, journal, twoside]{ieeetran} 
\usepackage{graphicx}
\usepackage{xcolor}
\usepackage[ruled, vlined]{algorithm2e}
\usepackage{lipsum}
\usepackage[hidelinks]{hyperref}

\usepackage{bm}

\usepackage{amsmath}

\DeclareMathOperator*{\argminA}{arg\,min} 
\newcommand{\trsp}{{\scriptscriptstyle\top}}

\usepackage[
backend=bibtex,
doi=false,isbn=false,url=false,eprint=false,
maxcitenames=9,
maxbibnames=9,
labeldate=year,
giveninits=true,
]{biblatex}

\AtEveryBibitem{\clearfield{day}}

\author{Cem Bilaloglu, Tobias Löw, and Sylvain Calinon%
\thanks{Manuscript received: June, 20, 2023; Revised September, 20, 2023; Accepted October, 18, 2023.}
\thanks{This paper was recommended for publication by Editor L. Pallottino upon evaluation of the Associate Editor and Reviewers' comments. 
This work was supported by the State Secretariat for Education, Research and Innovation in Switzerland for participation in the European Commission's Horizon Europe Program through the INTELLIMAN project (\url{https://intelliman-project.eu/}, HORIZON-CL4-Digital-Emerging Grant 101070136) and the SESTOSENSO project (\url{http://sestosenso.eu/}, HORIZON-CL4-Digital-Emerging Grant 101070310).} 
\thanks{The authors are with the Idiap Research Institute, CH-1920 Martigny, Switzerland and also with the EPFL, 1015 Lausanne, Switzerland
        {\tt\footnotesize cem.bilaloglu@idiap.ch; tobias.loew@idiap.ch; sylvain.calinon@idiap.ch}}%
\thanks{Digital Object Identifier (DOI): see top of this page.}
}

\title{Whole-Body Ergodic Exploration with a Manipulator Using Diffusion}

\begin{document}

\maketitle

\input{sections/0Abstract}
\input{sections/1Introduction}
\input{sections/2RelatedWork}
\input{sections/3Method}
\input{sections/4Experiments}

\input{sections/5Conclusion}

\printbibliography

\end{document}

%% file: sections/0Abstract.tex
\begin{abstract}
This paper presents a whole-body robot control method for exploring and probing a given region of interest. The ergodic control formalism behind such an exploration behavior consists of matching the time-averaged statistics of a robot trajectory with the spatial statistics of the target distribution. Most existing ergodic control approaches assume the robots/sensors as individual point agents moving in space. We introduce an approach that decomposes the whole-body of a robotic manipulator into multiple kinematically constrained agents. Then, we generate control actions by calculating a consensus among the agents. To do so, we use an ergodic control formulation called heat equation-driven area coverage (HEDAC) and slow the diffusion using the non-stationary heat equation. Our approach extends HEDAC to applications where robots have multiple sensors on the whole-body (such as tactile skin) and use all sensors to optimally explore the given region. We show that our approach increases the exploration performance in terms of ergodicity and scales well to real-world problems. We compare our method in kinematic simulations with the state-of-the-art and demonstrate the applicability of an online exploration task with a 7-axis Franka Emika robot. Additional material available at \noindent{\url{https://sites.google.com/view/w-ee-d/}}
\end{abstract}

%% file: sections/1Introduction.tex
\section{Introduction}
\IEEEPARstart{V}{arious} exploration tasks require physical interaction to collect information due to contact requirements or sensory occlusion. 
Existing work in tactile exploration ranges from object shape reconstruction with tactile skins \cite{bauzaTactileMappingLocalization2019}, and
probing for stiffness mapping
\cite{ayvaliUsingBayesianOptimization2016a}, to exploring wrench space of an articulated object \cite{löwEfficientRepresentationLearning} and exploring end-effector poses for insertion tasks \cite{shettyErgodicExplorationUsing2022}. Although it is possible to formulate these tasks as an optimization of information measures \cite{taylorActiveLearningRobotics2021}, setting up the objective is challenging and computation is expensive. Fortunately, for a subset of problems, one can formulate the autonomous exploration as a coverage of a region (target distribution).

Existing work in spatial coverage focuses on multi-agent systems with high-range sensors mounted on drones and ground robots (camera, LiDAR, time of flight sensor arrays \cite{bilalogluNovelTimeofFlightRange2022}) for increasing coverage speed. Applications of these methods explore vast regions such as crop fields \cite{ivicAutonomousControlMultiagent2019} or bridges \cite{ivicMultiUAVTrajectoryPlanning2023}. However, robotic manipulators are advantageous for exploring small but complex target distributions with physical interaction. Nevertheless, existing methods do not consider the link footprints and the kinematic chain composing a robot manipulator’s whole-body for exploration and lack the potential performance gains of increased sensor coverage. We argue that, as the accessibility of anthropomorphic robotic hands \cite{shorthoseDesign3DPrintedSoft2022}, and arms equipped with joint force-torque sensors and tactile skins \cite{albiniExploitingDistributedTactile2021,fanEnablingLowCostFull2022} increases, so will the need for whole-body exploration. Therefore, in this letter, we propose a whole-body ergodic exploration method for robotic manipulators that we summarized in Figure \ref{fig:summary}.
\input{floats/summary}

%% file: floats/summary.tex
\begin{figure}[]
        \centering
        \includegraphics[width=0.9\linewidth]{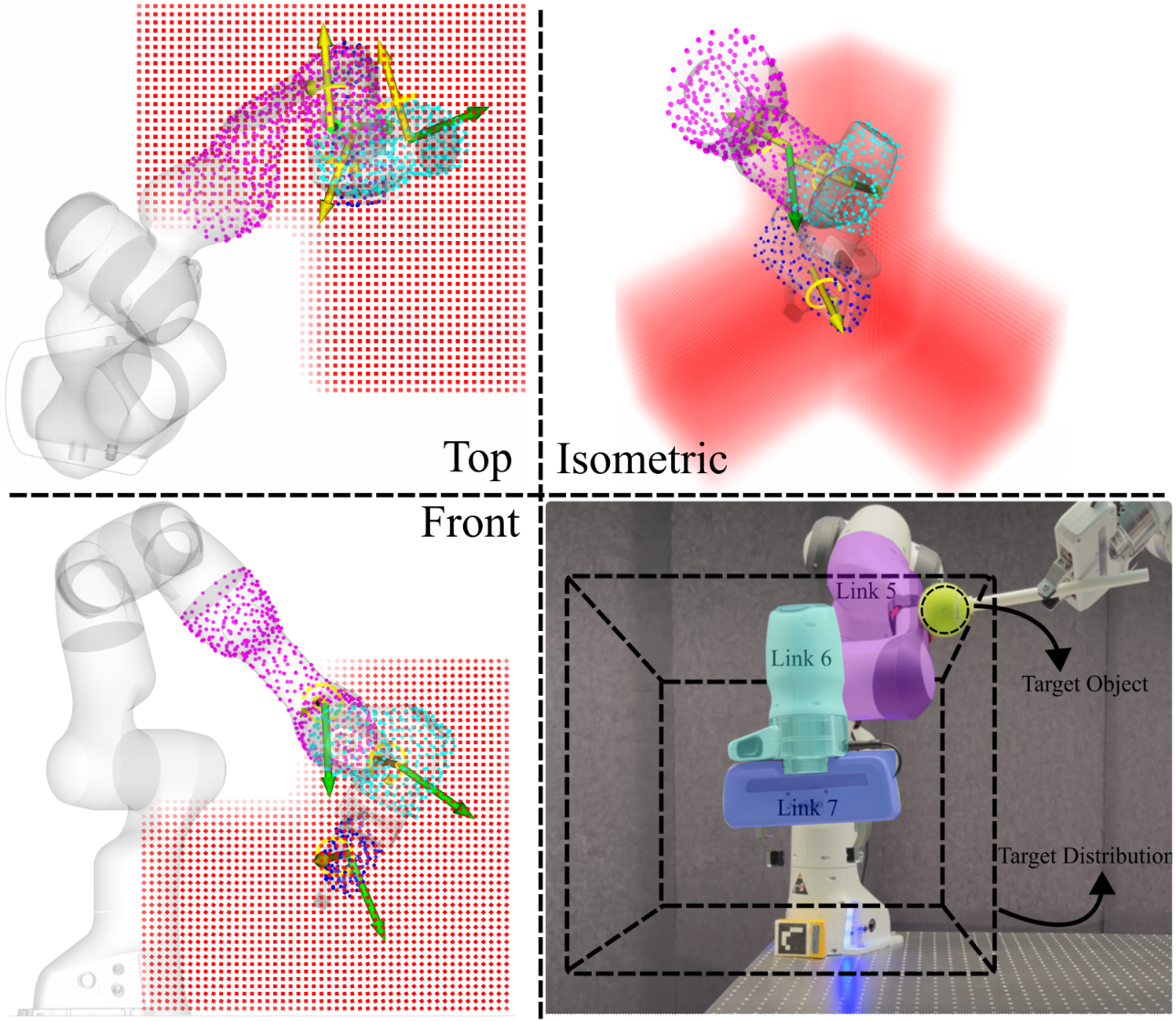}
        \caption{Whole-body exploration of a target distribution using the last three links of the robot manipulator. In kinematic simulation, the exploration target is given in red. In the real-world experiment, the robot explores the cube region in dashed lines to localize a target object (a tennis ball whose location is unknown). Blue, turquoise, and purple spheres are the virtual agents constrained to the $5$-th, $6$-th, and $7$-th links, respectively. The green and yellow arrows show the net virtual force and torque acting on each link's center of mass calculated by our agent weighing strategy. We further weigh the net wrenches acting on the active link to obtain the consensus control action for the robot.}
        \label{fig:summary}
\end{figure}

%% file: sections/2RelatedWork.tex
\section{Related Work}
The common challenge in autonomous exploration is the inherent curse of dimensionality. Indeed, a naive exploration based on random actions quickly becomes infeasible and instead requires intelligent strategies leveraging prior information \cite{ayvaliErgodicCoverageConstrained2017}. In settings with minimal uncertainty about the task priors, the best approach is to use coverage path planning (CPP) \cite{bircherRecedingHorizonPath2018, yoderAutonomousExplorationInfrastructure2016} or informative path planning (IPP) if the information is unevenly distributed \cite{popovicInformativePathPlanning2020,zhuOnlineInformativePath2021a}. Nevertheless, planning methods require specifying the horizon in advance and boil down to solving a trajectory optimization problem that is intractable for the most general case.

Control methods are robust to uncertainty compared to planning approaches and can be used when the terminal time for planning is unknown \cite{mathewMetricsErgodicityDesign2011}. In \cite{mathewSpectralMultiscaleCoverage2009a}, Mathew and Mezić proposed to use ergodicity---a concept that originated from statistical physics providing a metric that measures the difference between the target distribution and time-averaged statistics of agent trajectories \cite{abrahamErgodicMeasureActive2021} for control. This particular method called \emph{spectral multiscale coverage} (SMC), minimizes the ergodic metric by matching the Fourier series weights of the target distribution and the reconstructed distribution from the agent trajectories (coverage). Although the original formulation uses the Dirac delta function as the agent's footprint to simplify the computation of coverage, Ayvali \emph{et al.} \cite{ayvaliErgodicCoverageConstrained2017} proposed using KL-divergence for arbitrary agent footprints, which also alleviates the need for approximating the target distribution by a finite number of basis functions. However, the techniques based on KL-divergence are sampling-based planners, and they were not tested in online control settings. In \cite{shettyErgodicExplorationUsing2022}, Shetty \emph{et al.} proposed an online ergodic exploration technique for peg-in-hole tasks. They extended the SMC algorithm by a low-rank approximation called tensor train factorization to scale the computation of Fourier series weights in the 6-D pose space describing the end-effector location.

A recent alternative to SMC-based methods is \emph{heat equation-driven area coverage} (HEDAC) \cite{ivicErgodicityBasedCooperativeMultiagent2017}. HEDAC encodes the target distribution as a virtual heat source and calculates the potential field resulting from diffusion, modeled as heat conduction in a uniform medium. Virtual heat conduction provides a model to smooth the gradient field and propagate information about unexplored regions to the agents. This technique is based on the diffusion (heat) equation, an extensively studied partial differential equation (PDE) that can be solved in various domains such as mesh surfaces point clouds and by using explicit or implicit time stepping schemes \cite{macdonaldImplicitClosestPoint2010a}. Additionally, it is possible to introduce internal and external domain boundaries where no heat conduction is allowed to encode exploration with embedded obstacle avoidance behavior. For instance, Ivi\'c \emph{et al.} adopted a finite element method to solve the heat equation on a planar domain with obstacles modeled as internal boundaries \cite{ivicConstrainedMultiagentErgodic2022} using the Neumann boundary conditions. They later extended this approach to a three-dimensional planning setting \cite{ivicMultiUAVTrajectoryPlanning2023} because the re-planning frequency was too low for online control. Existing work in HEDAC focuses on multi-agent systems, with the only exception being drozBot \cite{löwDrozBotUsingErgodic2022}, a robot manipulator drawing artistic portraits. DrozBot solves the non-stationary heat equation for a single explicit timestep to have an artistic effect similar to doodling. The method considers only a single point (the tip of the pen) for the planar coverage and not the whole-body of the manipulator.

In this work, on the other hand, we consider the robot's link footprints and the kinematic chain for increasing sensor coverage. For that purpose, we decompose the whole-body into kinematically constrained virtual agents. The primary challenge here is to formulate a locally consistent exploration behavior in time and space, combining agents for global exploration with the robot. We achieve this by solving the non-stationary diffusion equation to slow the diffusion to increase local exploration. Then, we combine local exploration and introduce weighting strategies on the agent and link levels to simplify reaching a consensus among the agents. Consequently, we present the first whole-body ergodic exploration method and the first three-dimensional control implementation of the HEDAC approach. We summarize our contributions as follows:
\begin{itemize}

\item Increased sensor coverage by modeling the whole-body as a collection of virtual exploration agents constrained to the robot’s kinematic chain;

\item Formulating a locally consistent exploration behavior in time and space, combining agents for global exploration;

\item Controlling the robot with consensus among virtual agents and links by the introduced weighting strategies

\end{itemize}

%% file: sections/3Method.tex
\section{Whole-Body Control using Kinematically Constrained Virtual Agents}
\input{floats/hedac_explain}

\subsection{Smooth Potential Field Resulting from Diffusion}
We extend the state-of-the-art ergodic control technique HEDAC to obtain the potential field guiding the exploration behavior. Original implementation \cite{ivicErgodicityBasedCooperativeMultiagent2017} solves stationary ($ \dot{u}(\bm{x},t) = 0$) diffusion with virtual source $s(\mathbf{x}, t)$, sink $a(\mathbf{x}, t)$ and convective loss $u(\mathbf{x}, t)$ terms
\begin{equation}
    \alpha \cdot \Delta u(\mathbf{x}, t)=\beta \cdot u(\mathbf{x}, t)+\gamma \cdot a(\mathbf{x}, t)-s(\mathbf{x}, t).
    \label{eq:hedac}
\end{equation}
We summarize the HEDAC algorithm which uses the potential field $u(\mathbf{x}, t)$ resulting from integrating \eqref{eq:hedac} in Figure \ref{fig:hedac_explain}. We modify HEDAC to formulate a locally consistent exploration behavior in time and space, combining agents for global exploration. For that purpose, unlike the original formulation, we solve non-stationary ($ \dot{u}(\bm{x},t) \neq 0$) diffusion to increase local exploration by slowing down the diffusion
\begin{equation}
 \dot{u}(\bm{x},t)=\alpha \cdot \Delta u(\bm{x},t)- u(\mathbf{x}, t) + s(\bm{x},t).
     \label{eq:heat}
\end{equation}
We omit the sink term because we want to combine agents therefore inter-agent collision is not a concern. Moreover, we use a single parameter ($\alpha$) for tuning the exploration behavior as proposed in \cite{ivicMultiUAVTrajectoryPlanning2023}. We numerically integrate \eqref{eq:heat} using an explicit time-stepping scheme
\begin{equation}
 u(\bm{x},t+1)= u(\bm{x},t) +  \dot{u}(\bm{x},t) \cdot \delta t,
     \label{eq:integration}
\end{equation}
where the maximum stable timestep $\delta t$ is a function of the diffusion rate $\alpha$ and given by Courant–Friedrichs–Lewy condition
\begin{equation}
    \delta t\left(\sum_{i=1}^{N_d} \frac{\alpha_i}{\delta x_i}\right) \leq 1,
    \label{eq:cfl}
\end{equation}
with the spatial discretization $\delta x$ and the number of spatial dimensions $N_d$. We introduce the parameter $N_k$, for the number of integration steps we take using the maximum stepsize $\delta t$ computed by \eqref{eq:cfl}. Note that choosing high values for $N_k$ would mean integrating \eqref{eq:heat} until $u(\bm{x},t+1)-u(\bm{x},t) \approx 0$ and would be equivalent to stationary diffusion.

\subsection{Kinematically Constrained Virtual Agents}
We define \emph{virtual agents} as atomic particles that compose a rigid body and interact with the potential field $u(\bm{x},t)$. Depending on the task, virtual agents abstract a sensor/tool used for physical interaction during  exploration. Additionally, we use the term \emph{whole-body} if the sensor/tool spans multiple bodies on different links of the manipulator. By construction, we can locate each agent in the potential field using the forward kinematics function $f_{\text{kin}}$ of the robot
\begin{equation}
    \bm{x}_{i,j} = \bm{f}_{\text{kin}}(\bm{q},i,j)\quad \forall i=1,\ldots, N_j \quad \forall j=1,\ldots, M,
    \label{eq:fkin}
\end{equation}
where $N_j$ is the number of virtual agents on the j-th link, and $M$ is the number of links composing the whole-body. $\bm{x}_{i,j}$ is the position of the $i$-th agent on the $j$-th link and $\bm{q}$ is the vector of joint variables.

\subsection{Active Agents and Local Weighting}

We call the virtual agents that we use for computing coverage but not contributing to the control action as \emph{passive}. We use the term \emph{passive} because these agents explore regions indirectly as a secondary effect of the robot's primary goal. Still considering their coverage relieves the need to revisit those locations later. On the other hand, \emph{active} agents, contribute to the control command of the robot with their local exploration goal in addition to coverage.

According to our model, the potential field exerts a fictitious force on each active agent based on the gradient of the potential field, and we multiply this force by an agent weight $w_{i,j}$
\begin{equation}
   \bm{f}_{i,j} = w_{i,j}  \nabla u(\bm{x}_{i,j}(t),t).
   \label{eq:force}
\end{equation}

The naive method of computing the agent weights is to use a uniform weighting strategy, thus assigning equal importance to every agent. However, this is suboptimal since the significance of agents differs depending on their position in the potential field and the current state of the potential field itself. The value of the potential field at a given point encodes how much this particular region is underexplored. Accordingly, we embed this information by using the local value sensed by the agent as its weight. Hence, agents on the frontier of exploration (the ones closer to the underexplored regions) will have a higher weight than those on the overexplored regions. Thus, we set the normalized weight of the $i$-th agent on the $j$-th link as
\begin{equation}
    w_{i,j} = \frac{\tilde{w}_{i,j}}{\sum_{i=1}^{N_j} \tilde{w}_{i,j}}
    \quad \text{with} \quad
        \tilde{w}_{i,j} = u(\bm{x}_{i,j}(t),t),
    \label{eq:weight}
\end{equation}
and call it \emph{local weighting strategy}. Note that, the local weight is a function of the potential field, i.e. both space and time, and is therefore computed online. We show the difference between local and uniform weighting strategies in Figure \ref{fig:weighting}.
\input{floats/weighting}

\subsection{Active Links and Manipulability Weighting}
Similarly to active agents, we call a rigid body composed of active agents \emph{active link} and compute the net force and moment acting on it by all the agents
\begin{equation}
    \bm{f}_{\text{net}_j} = \sum_{i=1}^{N_j}\bm{f}_{i,j}, \quad
        \bm{m}_{\text{net}_j} = \sum_{i=1}^{N_j} \bm{r}_{i,j} \times \bm{f}_{i,j},
\end{equation}
where $\bm{r}_{i,j}$ is the displacement vector connecting the active agent and the $j$-th link's center of mass. We concatenate force and moment into a net wrench acting on the $j$-th link of the manipulator. For the simplest kinematic control strategy, we set the desired twist of the link $\bm{v}_{\text{des}_j}$ equal to the net wrench acting on the $j$-th link 
\begin{equation}
        \bm{v}_{\text{des}_j} = \left[
    {\begin{array}{cc}
        \bm{f}_{\text{net}_j}^\trsp & \bm{m}_{\text{net}_j}^\trsp
    \end{array}}
    \right]^\trsp
\label{eq:twist}
\end{equation}
corresponding to having identity inertia.

Here the twist commands that we generate correspond to consensus among the agents because the gradient field exerts similar forces to neighboring agents (local consistency) since we set the local cooling term used for collision avoidance in HEDAC to zero and slowed down diffusion by solving the non-stationary diffusion equation. Moreover, the gradient (force) field resulting from spatial diffusion is smooth, and we perform first local exploration (moving based on the force field) and then global exploration (by the propagation of information as diffusion).

We weight the contribution of the active links to the control command because each link has different \emph{manipulability} \cite{yoshikawaManipulabilityRoboticMechanisms1985} and volume $\nu$, resulting in distinct volumetric coverage rates. For that purpose, we compute the link weights using the scalar manipulability index $\mu$
\begin{align}
    w_j&= \nu_j \mu_j,\\
    \text{with} \quad \mu_j &= \sqrt{\det\big(\bm{J}_j(\bm{q})\bm{J}_j(\bm{q})^{\trsp}\big)},
    \label{eq:link_weight}
\end{align}

where $\bm{J}_j(\bm{q})$ is the Jacobian of the $j$-th active link computed at its center of mass.

\subsection{Consensus Control for Whole-Body}
Our setting consists of multiple weight-prioritized tasks, encoded as task velocities with corresponding Jacobians, and we would like to perform them in the least square optimal sense by exploiting the redundancy of our robot
\begin{align}
    \bm{\dot{q}}_{\text{des}}&= \argminA_{\dot{\bm{q}}}
    \;(\bar{\bm{v}}_{\text{des}}-\bar{\bm{J}}\dot{\bm{q}})^\trsp \bm{W} (\bar{\bm{v}}_{\text{des}}-\bar{\bm{J}}\dot{\bm{q}})\\
    \quad \text{with} \quad
\bm{W} &= \text{blockdiag}\big(\bm{I}w_1,\bm{I}w_2,\hdots,\bm{I}w_m)
\label{eq:optimal_weight}
\end{align}
where $\bar{\bm{J}}$ is the vertical concatenation of the Jacobians, and $\bar{\bm{v}}_{\text{des}}$ is the vertical concatenation of the task velocities of active links. We enforce task priorities by using the weighted pseudoinverse and computing the desired joint velocities as
\begin{equation}
    \dot{\bm{q}}_{\text{des}}= 
\left ( \bar{\bm{J}}^\trsp \bm{W} \bar{\bm{J}}\right)^{-1} \bar{\bm{J}}^\trsp \bm{W}
    \bar{\bm{v}}_{\text{des}}
    \label{eq:q_des}.
\end{equation}

Next, we use desired joint velocity to either kinematically simulate the robot $\bm{q}_{t+1} = \bm{q}_t + \dot{\bm{q}}_{\text{des}} \cdot \Delta t$ or as a desired joint velocity to an impedance controller. Then, we clamp the desired joint positions as the simplest strategy to comply with joint limits. We give the full procedure for robot control in Algorithm \ref{alg:main_loop}.

\input{floats/algorithm}

%% file: floats/hedac_explain.tex
\begin{figure}[]
        \centering
        \includegraphics[width=0.8\linewidth]{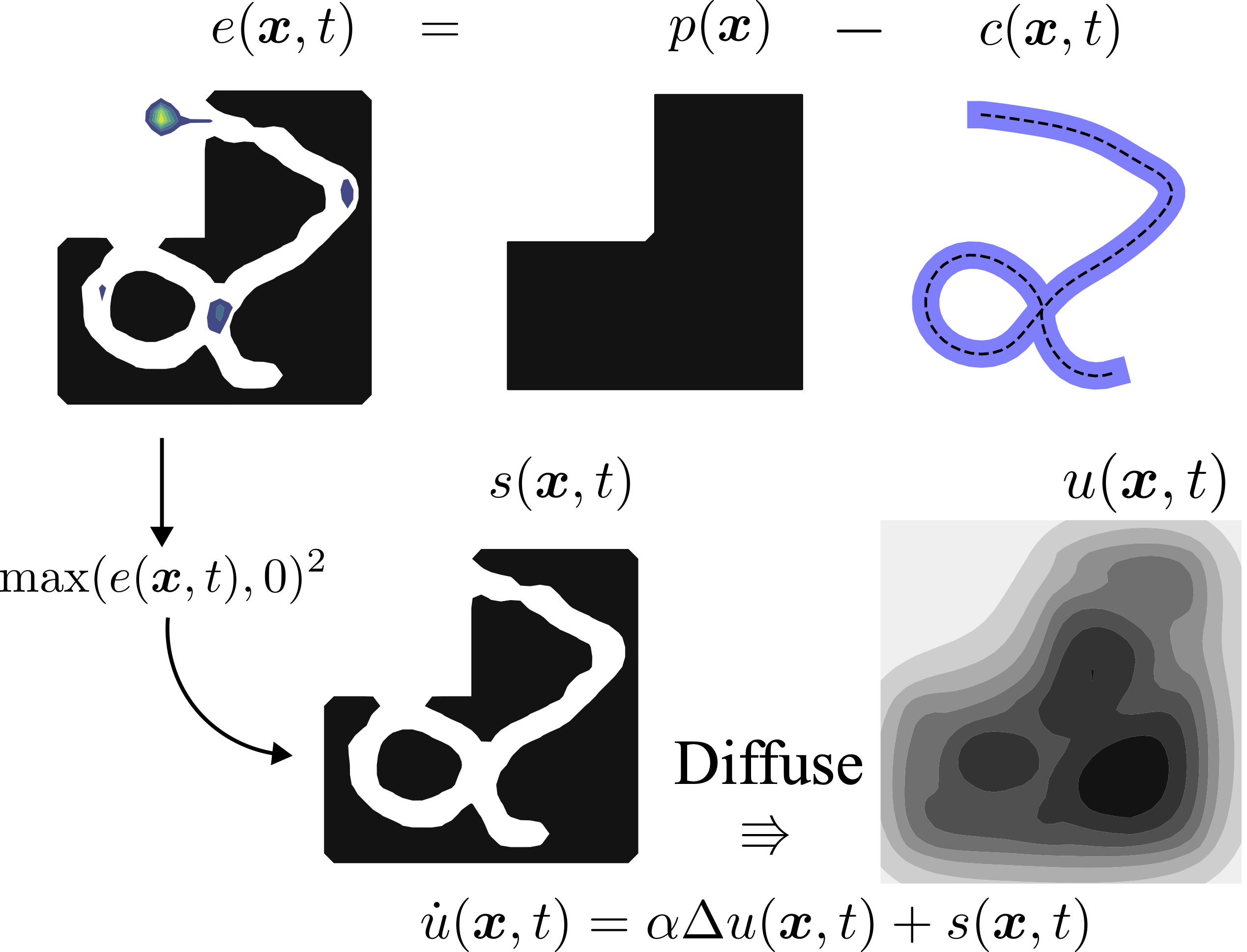}
        \caption{ The HEDAC 
        \cite{ivicErgodicityBasedCooperativeMultiagent2017} method computes the potential field $u(\bm{x},t)$ that guides the agents for ergodic exploration. The time-averaged coverage of the agent(s) $c(\bm{x},t)$ at time $t$ is subtracted from the target distribution $p(\bm{x})$ and positive values corresponding to unexplored regions are squared and used as the virtual heat source $s(\bm{x},t)$. The diffusion (heat) equation \eqref{eq:heat} is then used for diffusing the potential field and propagating information of unexplored regions to the agents.}
        \label{fig:hedac_explain}
\end{figure}

%% file: floats/weighting.tex
\begin{figure}[]
        \centering
        \includegraphics[width=0.7\linewidth]{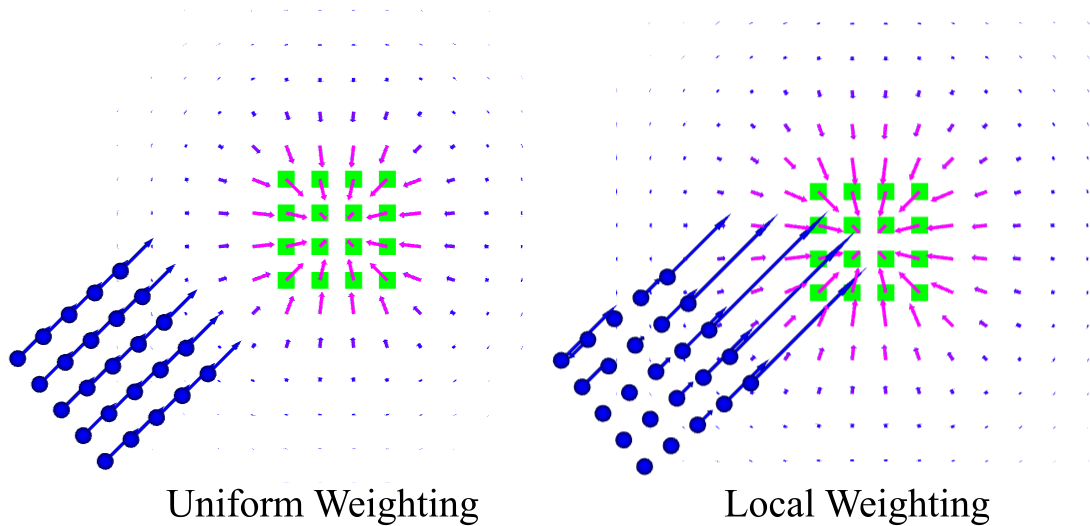}
        \caption{Comparison of uniform and local temperature weighting. The green square is the exploration target and small arrows show the temperature gradient. Blue dots and arrows show active agents and the force exerted on each agent after weighting.}
        \label{fig:weighting}
\end{figure}

%% file: floats/algorithm.tex
\begin{algorithm}
            \SetKwBlock{Empty}{}{}
            \KwIn{target distribution $p(\bm{x})$, forward kinematics function $\bm{f}_{\text{kin}}(\bm{q},i,j)$, initial joint configuration $\bm{q}_0$, number of active links $\bm{M}$, number of active agents $\bm{N_j}$ in $j$-th link}
            \KwOut{control commands $\bm{u}(t)$}
     
        initialize agents and bodies using using \eqref{eq:fkin} and $\bm{q}_0$
        
        compute $\delta t$ using \eqref{eq:cfl}
        
{
        \While{true} 
        {
            compute $c(\bm{x},t)$ and $s(\bm{x},t)$ as in \cite{ivicErgodicityBasedCooperativeMultiagent2017}

            \For{$k \gets 1$ to $N_k$}{
                integrate \eqref{eq:heat} using \eqref{eq:integration}
            }

            \For{all $M$ bodies}{

            \For{all $N_j$ agents}{
                        compute $\bm{x}_{i,j}$ and $\bm{f}_{i,j}$ using \eqref{eq:fkin} and \eqref{eq:force}
                }   
                
                compute $\bm{v}_{\text{des}_j}$ and $w_j$ using \eqref{eq:twist} and \eqref{eq:link_weight}
            }

            compute $\dot{\bm{q}}_{\text{des}}$ using  \eqref{eq:q_des}}}

        \caption{Whole-Body Exploration}
        \label{alg:main_loop}
    \end{algorithm}

%% file: sections/4Experiments.tex
\section{Whole-Body Exploration}
We give the videos for all the simulated and real-world experiments on the website.
\subsection{Simulated Experiments}
We performed kinematic simulations to measure the exploration performance. We used the normalized ergodicity over the target distribution as the exploration metric
\begin{equation}
    \varepsilon = \frac{\|\max\left(e(\boldsymbol{x},t),0\right)\|_2}{\int_{\Omega}p(\bm{x})d\bm{x}},
    \label{eq:ergodicity}
\end{equation}
where $p(\bm{x})$ is the target distribution, $e(\boldsymbol{x},t)$ is the residual given by $p(\bm{x})-c(\boldsymbol{x},t)$ as given in Figure \ref{fig:hedac_explain} and calculated as in \cite{ivicErgodicityBasedCooperativeMultiagent2017}. This metric shows how good the time-averaged statistics of agent trajectories match the target distribution, and lower values indicate higher performance. 

\subsubsection{Planar Experiments}
\input{floats/x_trajectory}
We first present the results of planar simulations showing the exploration performance qualitatively using images. We tested four different configurations: (i) \emph{SMC}, (ii) \emph{passive}, (iii) \emph{active/stationary} using \eqref{eq:hedac}, and (iv) \emph{active/non-stationary} ($N_k=1$) using \eqref{eq:heat}. SMC and passive configurations use a single active agent at the tip of the last link that we control by the state-of-the-art ergodic control methods SMC and HEDAC, respectively, where the passive configuration also considers the coverage of the last link indirectly for the control. We used active agents sampled on the last link equally spaced by the unit distance for the active configurations. We computed the metric \eqref{eq:ergodicity} using the coverage of the last link for all of these configurations. We used a $75\times75$ grid for discretizing the exploration domain and $20$ basis functions for the SMC method. We plotted the trajectories and the target distribution in Figure \ref{fig:x_trajectory}. Next, we repeated the same experiment for two different target distributions starting from $100$ uniformly sampled initial joint configurations. We chose one diffuse and one fine-detailed target distribution to test the scenarios advantaging stationary/non-stationary diffusion, respectively. Although the exploration behavior would continue indefinitely, we stopped the simulation after $1000$ timesteps. We calculated the mean and standard deviation of the normalized ergodic metric and plotted the results in Figure \ref{fig:stacked_shaded}.

\input{floats/stacked_shaded}

\subsubsection{Three Dimensional Experiments}
In the first 3D experiment, we measured the effect of using multiple active links on the exploration performance. We assumed no prior for the exploration target and used a cube discretized on a $50\times50\times50$ grid where each point corresponds to $1$ cm as the target distribution. We placed the target in front of a 7-axis Franka Emika robot and sampled active agents on links 5, 6, and 7 using Poisson-disk sampling. Note that we used the same agent configuration for computing the coverage and the metric using \eqref{eq:ergodicity}, but we changed the number of active links used for computing robot control commands. Similarly to the planar experiments, we used 30 pre-sampled initial joint configurations and plotted the results in Figure \ref{fig:cube_shaded}. 
\input{floats/cube_shaded_box}
Secondly, we compared the exploration performance of the proposed method to that of using a search pattern. In this task, the robot explored the same target region using its whole-body composed of links 5, 6, and 7 until one of the links touched the sphere target placed at an unknown position. We repeated the experiment 30 times for randomly sampled target sphere positions but used fixed initial configurations corresponding to the end-effector poses 1, 2 given in Figure \ref{fig:cube_box}b. We recorded the time until the first contact and plotted the results in Figure \ref{fig:cube_box}a.
\input{floats/box}

\subsection{Real-world Experiment}
For the real-world experiments, we used a 7-axis Franka Emika robot with the same object localization task performed in kinematic simulation. We placed a tennis ball inside the target distribution as the target object at an unknown location. We first recorded the trajectory resulting from the exploration behavior and then tracked the trajectory with a stiff impedance controller to ensure that the robot safely contacted the tennis ball and did not hit the second robot or the stick. We ran the experiment until one of the links made contact with the target object, which we detected by using the joint torque sensors of the robot. We provide the experiment setup in Figure \ref{fig:real_world_experiment}.

\input{floats/real_world_experiment}

%% file: floats/x_trajectory.tex
\begin{figure*}[]
        \centering
        \includegraphics[width=1\linewidth]{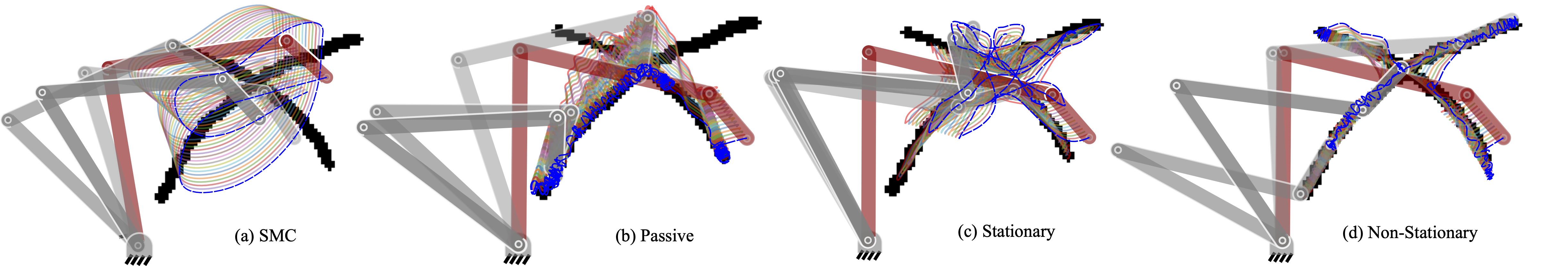}
        \caption{Planar exploration using different configurations. The black shape is the target distribution, and the colored lines are agent trajectories, where the blue dashed lines correspond to the agent's path at the tip of the last link. We show the configuration of the planar manipulator at equally spaced timesteps. Red is the start configuration, and the transparency decreases as time increases.}
        \label{fig:x_trajectory}
\end{figure*}

%% file: floats/stacked_shaded.tex
\begin{figure}[]
        \centering
        \includegraphics[width=0.9\linewidth]{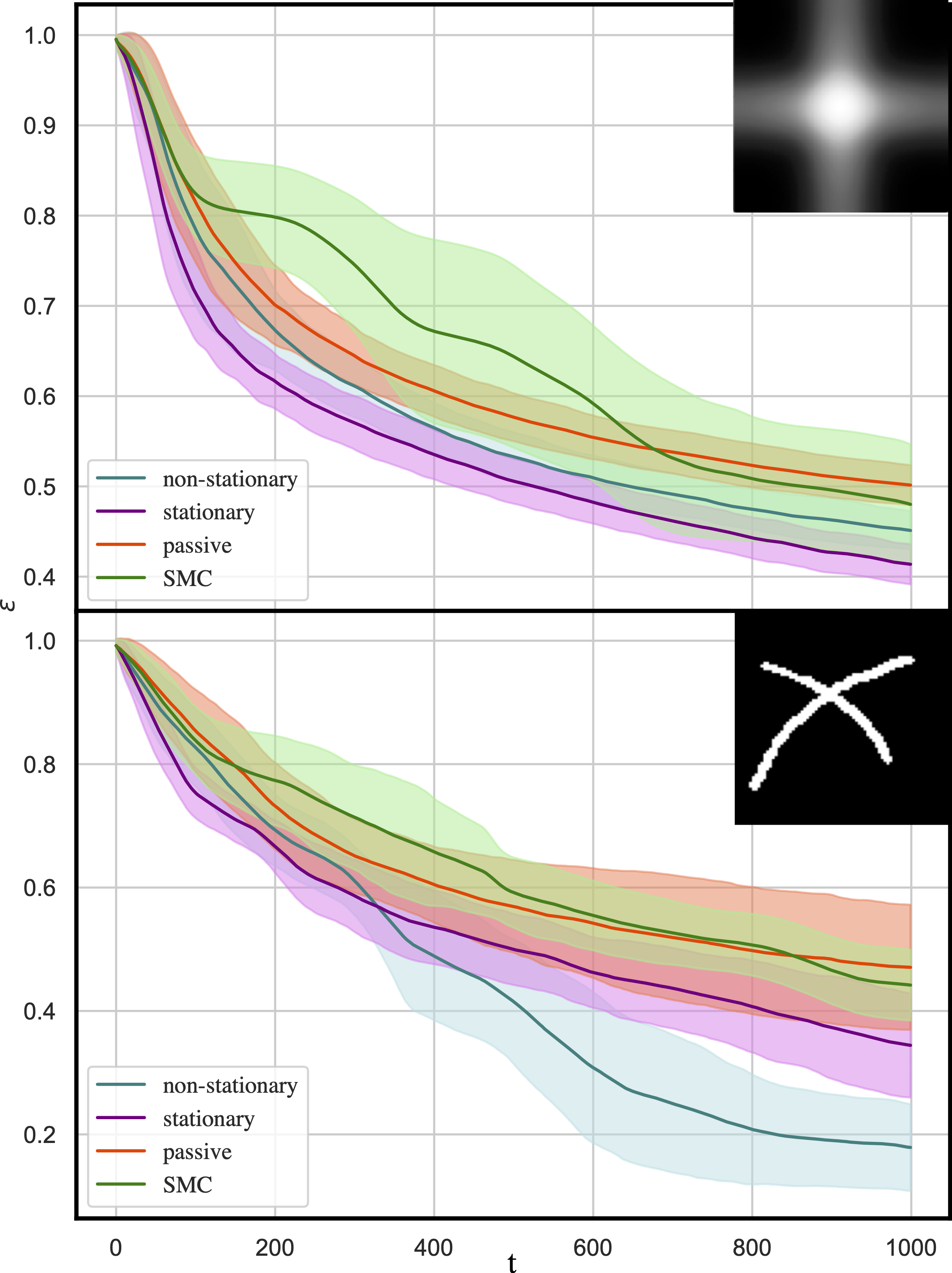}
        \caption{Coverage performance given by the normalized ergodic metric for different virtual agent configurations. Target distributions for the coverage task are given on the top right. }
        \label{fig:stacked_shaded}
\end{figure}

%% file: floats/cube_shaded_box.tex
\begin{figure}[]
        \centering
        \includegraphics[width=0.9\linewidth]{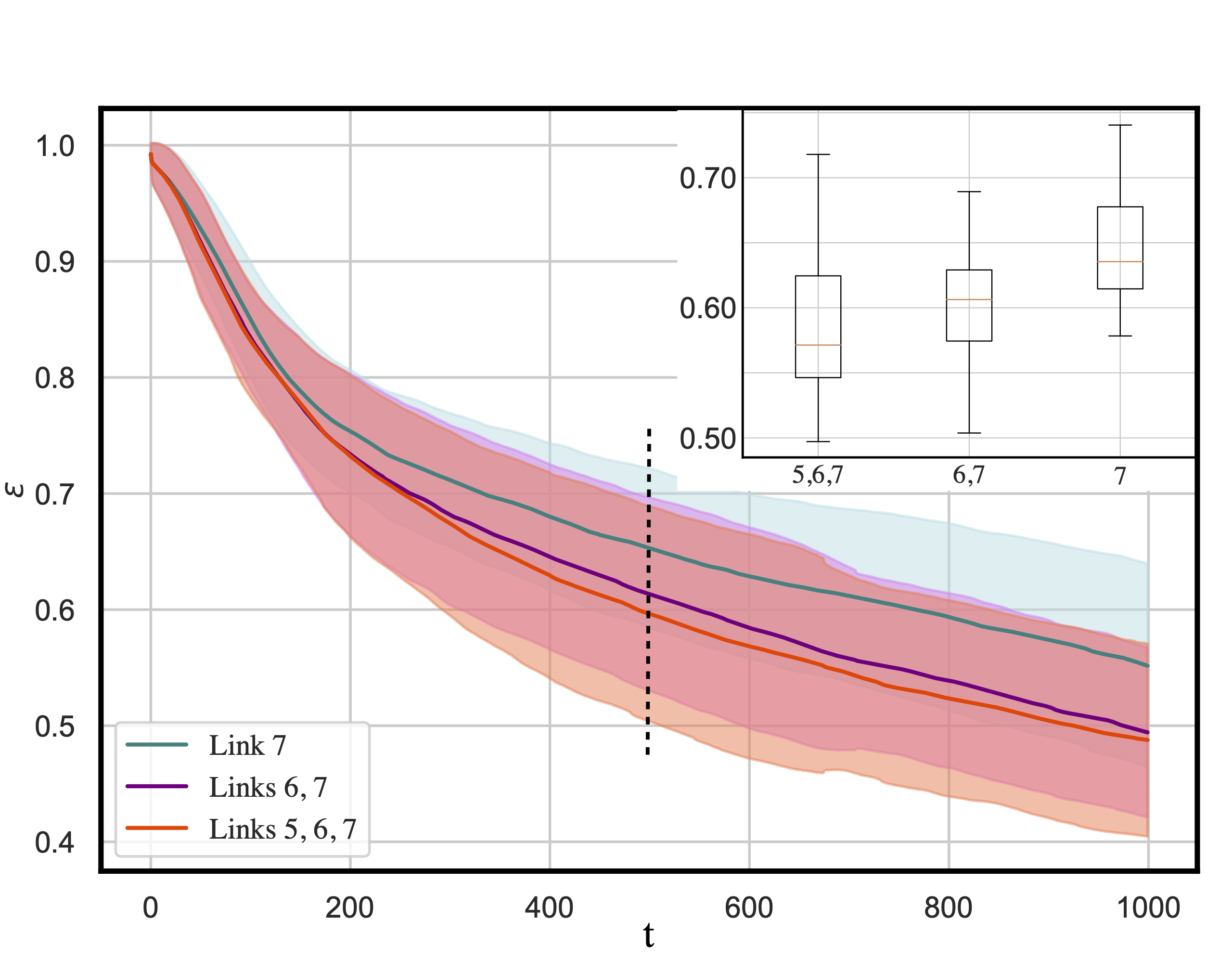}
        \caption{
        Coverage performance given by the normalized ergodic metric for different virtual agent configurations. The inset boxplot shows the performance in detail for each configuration at 500-th timestep. During the experiments, we set $N_k=3$ as an empirically found moderate value and attained a control frequency of $35$, $18$, and $13$ Hz for active agents on the last one, two, and three links respectively using a laptop processor.}
        \label{fig:cube_shaded}
\end{figure}

%% file: floats/box.tex
\begin{figure}[]
        \centering
        \includegraphics[width=0.8\linewidth]{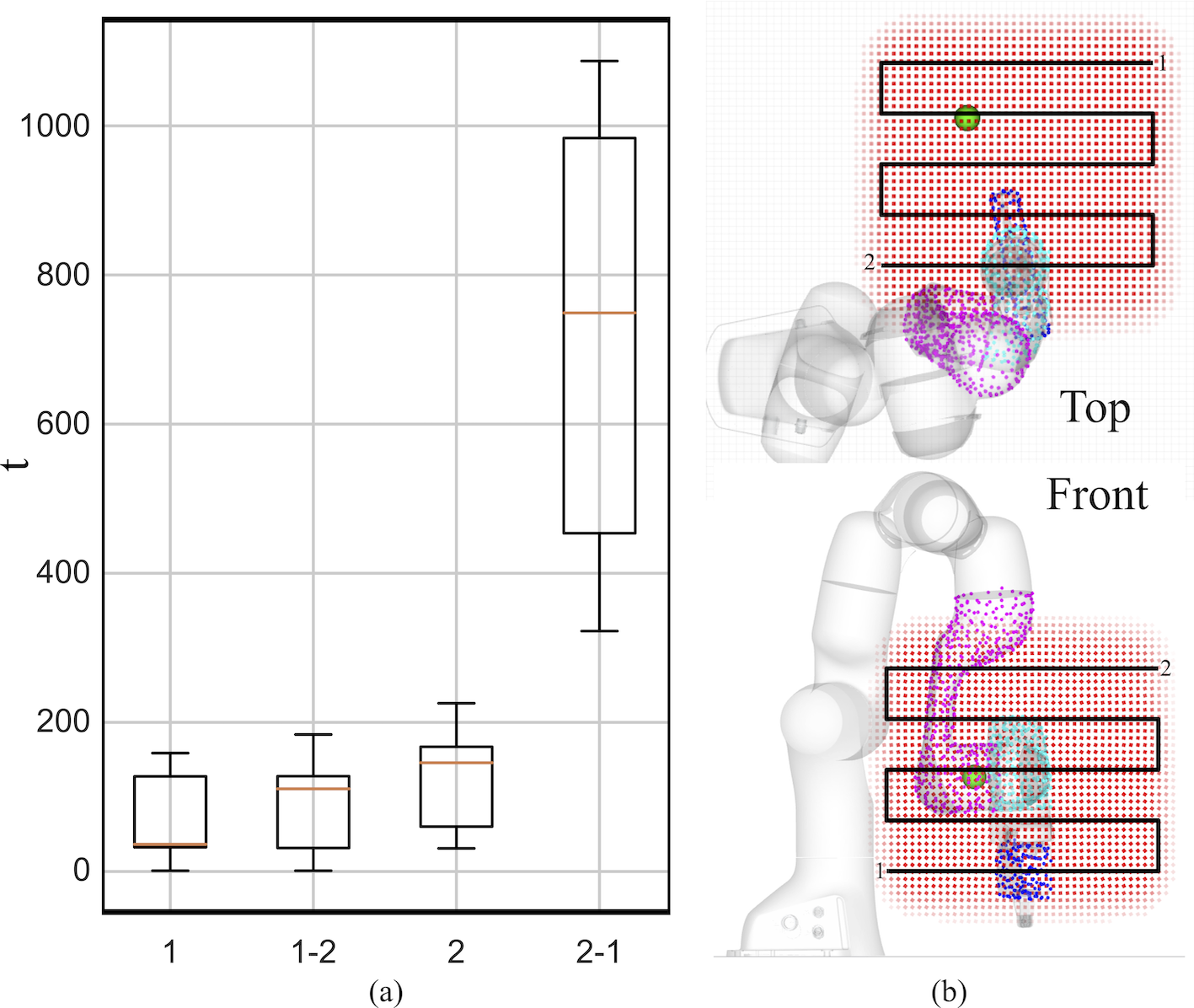}
        \caption{Coverage performance given by the timesteps required until the first contact of the whole-body with the target object (green sphere). a) The horizontal axis of the boxplot corresponds to our method and a planned search pattern starting from the initial end-effector poses 1 and 2. b) We show the initial poses 1, 2 and the direction of the search pattern is indicated by the ordering of the poses.}
        \label{fig:cube_box}
\end{figure}

%% file: floats/real_world_experiment.tex
\begin{figure}[]
        \centering
        \includegraphics[width=0.9\linewidth]{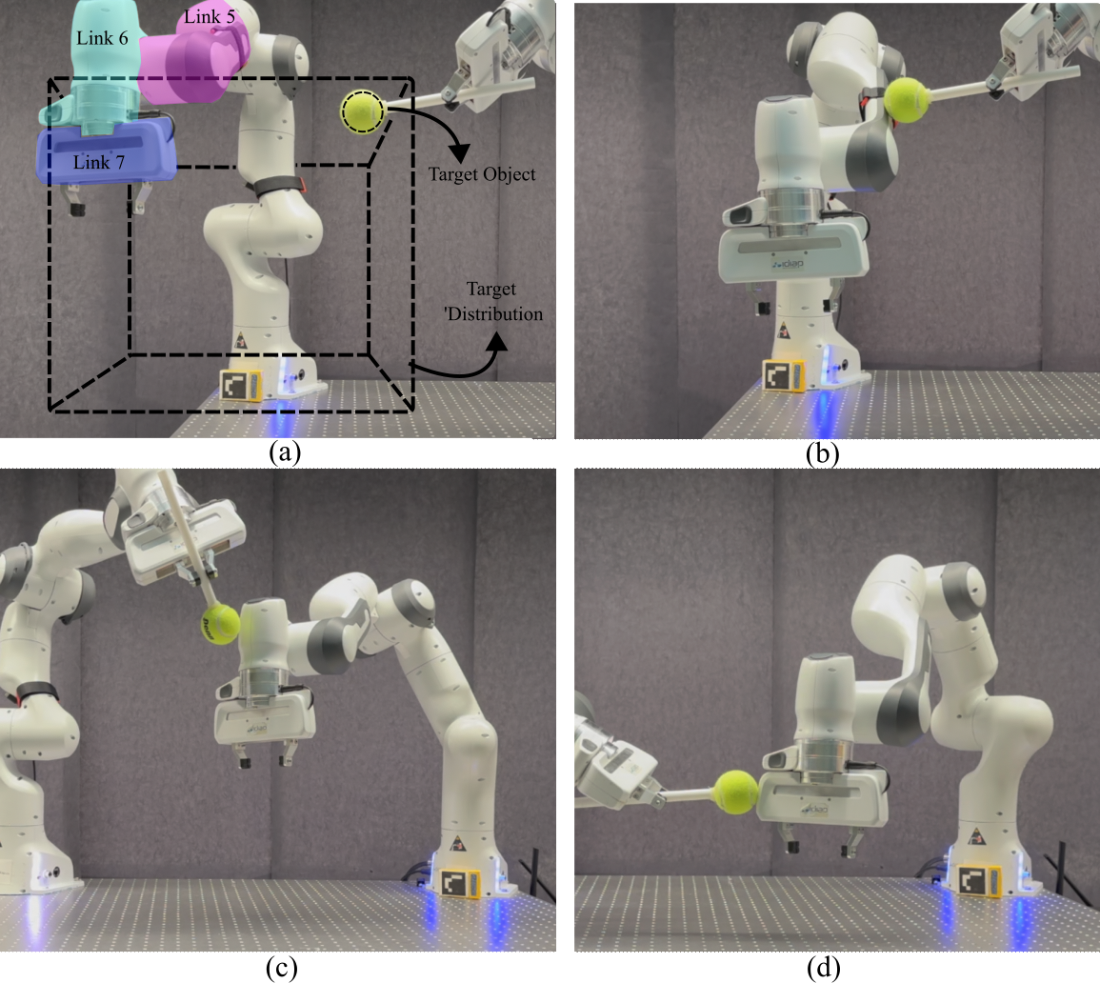}
        \caption{Real-world experiment of the robot exploring the cube in dashed lines using its last three links, until either b) link $5$, c) link $6$ or d) link $7$ contacts the target object.}
        \label{fig:real_world_experiment}
\end{figure}

%% file: sections/5Conclusion.tex
\section{Discussion}

We investigated the effect of using active/passive agents and stationary/non-stationary diffusion on the exploration trajectory in Figure \ref{fig:x_trajectory}. We see SMC and passive configurations do not, and stationary diffusion configuration can not efficiently align with the target. On the other hand, in the non-stationary case, agents align with the target distribution for most of their trajectories (only get misaligned when rotating for re-alignment). This is expected since non-stationary diffusion increases local exploration and agent coordination, resulting in better alignment behavior. Next, we tested the time evolution of the ergodic metric in Figure \ref{fig:stacked_shaded}. On top of the qualitative results of Figure \ref{fig:x_trajectory}, we can quantitatively argue active agents have higher performance than the state-of-the-art SMC method and the passive configuration. However, the performance difference between stationary/non-stationary diffusion depends on exploration time and the target distribution. This is because stationary diffusion ($N_k \gg 1$) smooths the details and coarsely explores diffuse targets but can not explore distal and fine-detailed regions. In contrast, non-stationary diffusion with $N_k=1$ performs significantly better in distal and fine-grained regions but slightly worse for a diffuse target. For $N_k=(1,10]$, we obtained results that interpolate the stationary/non-stationary diffusion results given in Figure 5. Accordingly, for the number of integration steps $N_k$, we recommend using a single step ($N_k=1$) for concentrated targets,  high values ($N_k\approx10$) for diffuse targets, and low to medium values ($N_k\approx3$) for generic cases.

In the first 3-D experiment, we see that using multiple links increases the exploration performance as depicted in Figure \ref{fig:cube_shaded}. However, this gain diminishes as we move to the links with less manipulability. Links closer to the base, including link 5, have negligible weights when computing the robot control command due to the normalization. Consequently, they practically become passive, contributing to coverage but not to the robot control commands. This partly explains why we observe marginal gains after link 6. Nevertheless, we think the primary reason is that the target distribution is small compared to the link size. This decreases the benefit of increasing the sensor footprint by using more links. Unfortunately, we can not make the target distribution larger due to the manipulator's joint limits and fixed base. Nonetheless, if the link and target size ratio were lower, the performance would increase with the sensor footprint. Two such examples are using a mobile manipulator where we can increase the target size or using the whole-body of a robot hand connected to an arm where we can decrease the link size relative to the target distribution.

In Figure \ref{fig:cube_box}, we compare the performance of the proposed method to using a planned search pattern. On the path from 1 to 2, links 5, 6, and 7 initially stay inside the target distribution, whereas on the path from 2 to 1, all links explore only at the end of the path. Accordingly, these two paths correspond to the best and worst-case scenarios for exploration. The results show that our approach is more robust to the initial robot configuration and can effectively leverage the whole-body as it performs better than the planned path of 1-2. Although here we used a uniform distribution in the form of a cube, ergodic exploration methods such as ours generalize to arbitrary target distributions, whereas pattern-based or coverage, informative path-planning approaches are limited to uniform distributions or simple geometries.

We showcased the method’s applicability with a manipulator in a real-world experiment and presented the first 3D control implementation of the HEDAC approach. The control frequency of the method primarily depends on (i) integrating the diffusion using \eqref{eq:integration} and (ii) the number of active agents and links. Note that our choice of solving non-stationary diffusion with \eqref{eq:heat} and using explicit time-stepping results in higher computational efficiency for small $N_k$. In contrast, for the stationary diffusion equation \eqref{eq:hedac} using implicit time-stepping reduces to sparse matrix multiplication with pre-inverted system matrix and is more efficient than the iterative solution. To reduce the effect of the number of active links and agents on computation, we decomposed links to multiple agents instead of using a single agent with the shape of the link. Because coverage computation using a single agent with the link shape can not be parallelized and is computationally expensive due to including more zero entries in the convolution kernel. In the end, integrating the diffusion equation is the main bottleneck in our implementation, and increasing the number of agents has minimal effect. Still, considering that the exploration performance gain is minimal and the link weight is almost zero for link 5, we think it is better to consider the last two links for the whole-body exploration if the robot base is fixed.

An alternative use case for the presented method is using it as a long-horizon planner anticipating the joint limits instead of a myopic controller. Although this would improve the performance, it would be minimal because, thanks to the smooth potential field resulting from diffusion, it is unlikely to get stuck in a local minima due to joint limits. Still, we perceive the limited workspace due to the joint limits and the fixed base as the primary limitation to using whole-body exploration. Using a mobile manipulator would drastically increase whole-body exploration's performance because the link size target size ratio can be lower, and we would benefit from even less manipulable but large links. However, a limitation of the current approach in such a scenario would be the computational complexity of solving diffusion in a larger domain. Fortunately, unsupervised learning techniques for approximating diffusion, such as tensor train decomposition and physics-informed neural networks, seem promising. 

Finally, real-world experiments showed we can perform whole-body tactile exploration using off-the-shelf torque-controlled robots without additional sensors. This result provides insight into alternative platforms equipped with joint torque sensing and searching for contacts, such as legged robots and multi-fingered robot hands. For instance, legged robots can use our method when exploring ground contacts, and robot hands can use it for grasping where each finger moves until contacting the object of interest. Although we can sense contact by joint torques, localizing it on the robot surface is challenging and requires whole-body tactile sensors. Accordingly, in the future, we plan to use the method with recursive target distribution updates and tactile sensors to reconstruct the physical properties of the environment that can only be measured through contact, such as deformability or friction.

\section{Conclusion}
In this letter, we presented a robot control method for efficiently exploring a target distribution using a robotic manipulator's whole-body. Unlike existing approaches for multi-robot systems using independent agents, we used kinematically constrained agents on the links of a robotic manipulator to increase coverage. We introduced active agents and links together with weighting strategies considering the shape and kinematic chain of the whole-body for exploration. We combined the agents composing the whole-body for global exploration by formulating a locally consistent exploration behavior in time and space using non-stationary diffusion. Lastly, we measured the performance of our method in terms of ergodicity in kinematic simulations and demonstrated its applicability in physical scenarios using the 7-axis Franka Emika in an object localization task.